\documentclass{article}
\usepackage{ijcai23}

\usepackage{times}
\usepackage{soul}
\usepackage{url}
\usepackage[utf8]{inputenc}
\usepackage[small]{caption}
\usepackage{graphicx}
\usepackage{amsmath}
\usepackage{amsthm}
\usepackage{booktabs}
\usepackage{algorithm}
\usepackage{algorithmic}
\usepackage[switch]{lineno}

\usepackage{color}
\usepackage{amssymb}
\usepackage{multirow}
\usepackage{mathrsfs}
\usepackage{bm}
\usepackage{pifont}

\newcommand{\cmark}{\ding{51}}%
\renewcommand{\vec}[1]{\boldsymbol{#1}}

\title{A Review on Zeroing Neural Networks}

\author{
    Author Name
    \affiliations
    Affiliation
    \emails
    email@example.com
}

\author{
Chengze Jiang$^1$
\and
Jie Gui$^{*1,2}$\and
Long Jin$^{3}$\And
Shuai Li$^4$
\affiliations
$^1$School of Cyber Science and Engineering, Southeast University, Nanjing, China\\
$^2$Purple Mountain Laboratories, Nanjing, China\\
$^3$School of Information Science and Engineering, Lanzhou University, Lanzhou, China\\
$^4$College of Engineering, Swansea University, Swansea, UK
\emails
czjiang@ieee.org,
guijie@seu.edu.cn,
longjin@ieee.org,
shuai.li@swansea.ac.uk
}

\begin{document}

\maketitle

\begin{abstract}
Zeroing neural networks (ZNNs) have demonstrated outstanding performance on time-varying optimization and control problems. Nonetheless, few studies are committed to illustrating the relationship among different ZNNs and the derivation of them. Therefore, reviewing the advances for a systematical understanding of this field is desirable. This paper provides a survey of ZNNs' progress regarding implementing methods, analysis theory, and practical applications. 
\end{abstract}

\section{Introduction}
Zeroing neural networks (ZNNs) are highly-efficient models that recent research has shown their outstanding performance on dynamic optimization and control tasks. ZNNs are implemented by an error function defined according to the concerned problem rather than a norm-based energy function. Besides, ZNNs circumvent the lagging error by exploiting both present and previous data to predict future data. The effect of ZNNs is devoted to designing an evolution function with an effective activation function and scale parameter to zeroing the corresponding error function for solving dynamic problems in a predictable manner. Therefore, developing a relatively simple discrete-time method with a lower calculation time is essential. In general, methods for solving time-varying problems are effective enough to satisfy the requirement of real-time computation. \footnote{$^{*}$Corresponding author.}
\par
The key idea behind ZNNs is the construction of an error function based on the specific problem and the design of an appropriate evolution function for real-time revising the error function. The core components of ZNNs are the evolution formula, activation function, and scale parameters, which determine the intrinsic feature and convergence speed, respectively. According to these factors, ZNNs evolve different variants, and we revisit current advances in Section \ref{PVM}. Various ZNNs construction methods and model advances are presented in Section \ref{VM}. Then, discretization methods of ZNNs are summarized in Section \ref{MDM}. After that, comprehensive and systematic theoretical analyses are introduced for investigating and guiding the development of ZNNs in Section \ref{TAsection}. Besides, ZNNs have been applied in many areas involving robot kinematics, multi-objective optimization, image processing, etc. Finally, the applications of ZNNs are summarized in Section \ref{App}.
\subsubsection{Comparison Between this and Previous Reviews}
It has been reported that several papers devote to summarizing the advance of ZNNs. However, these papers either do not include recent progress \cite{JinSurvey} or are limited to a subfield. For example, the literature \cite{VPZNNOne} is focused on varying-parameter ZNNs, while \cite{PNNJin} commits to reviewing the projection neural networks. Therefore, it is necessary to revisit recent research trends of ZNNs to help the community better understand and benchmark the current advance. To this end, this work provides a comprehensive summary of ZNNs covering methodology, analysis theory, and application, especially recent progression.

\section{Overview of ZNNs}\label{PVM}
In this section, we introduce the formulation process of ZNNs and then lead to research topics in this field. Additionally, broad concerns or representative problems are introduced to intuitively illustrate the design, transformation, and establishment process of ZNNs.
\subsection{ZNN Solution Framework}
The design and implementation framework of ZNNs are presented as follows for outlining the key components and construction procedure to lay a foundation for further extension. 
\begin{itemize}
	\item Step 1: Define the error function $\vec{e}(t)$ according to the target problem to monitor and measure the residual error of the ZNN model.
	\item Step 2: Design and adjust the evolution formula $\vec{\dot e}(t)=F(\vec{e}(t))$, where the $F(\cdot)$ needs to be defined as it can make the ZNN model converge to a steady state over time (i.e., error function $\vec{e}(t)$ converges to zero).
	\item Step 3: Solving the corresponding implicit or explicit model $M(\vec{\dot e}(t),\vec{e}(t),t)$ obtained by combining error function $\vec{e}(t)$ and evolution formula $\vec{\dot e}(t)$ to extract the theoretical solution of the specific problem in real-time.
\end{itemize}
The evolution function is the cornerstone of ZNNs and determines the approach and performance of ZNNs to solve the problem. In other words, appropriate and efficient evolution functions can significantly improve the problem solving performance of ZNNs \cite{FTZNNBook}. Existing evolution functions are presented in Table \ref{EF}. Then, one natural question is whether it is better to enlarge the scale parameter, modify the evolution function, or both. For instance, the evolution formula of the original ZNN (OZNN) is presented as follows, and other models can be derived similarly:
\begin{linenomath*}
\begin{equation}\label{ZNNEF}
	\vec{\dot e}(t)=-\gamma \Psi(\vec{e}(t)),
\end{equation}
\end{linenomath*}
where the scale parameter $\gamma > 0$ is used to control the convergence speed and $\Psi(\cdot)$ represents activation function.
\par
The activation function is a research hotspot in the ZNNs field, whose embedded improvement can accelerate the convergence or pose extra characteristics \cite{LSComplex}. Essentially, the activation function usually accelerates convergence through the gain parameter without changing the sign of the parameter thereby influencing the scale parameter to achieve better convergence. Meanwhile, exploiting better scale parameters in the evolution function can improve the performance and simplify the tuning process \cite{ZJZOne}. Recent progress in this line of work focuses on developing more effective approaches for realizing time-varying or adaptive scale parameters instead of the fixed-valued scale parameter. These improvements can not only promote the convergence speed but also improve the robustness of the model under noise perturbed without additional structures or preprocess \cite{PTVP}.

\subsection{Problem Transformation and Extension}
In this part, a dynamic quadratic minimization (DQM) problem is employed to illustrate the transformation process when realizing ZNNs. First, the DQM problem is defined as
\begin{linenomath*}
\begin{equation}\label{DQM}
	\text{min}_{\vec{x(t)}}~\frac{1}{2}\vec{x}^{\text{T}}(t)A(t)\vec{x}(t)-\vec{b}^{\text{T}}(t)\vec{x}(t),
\end{equation}
\end{linenomath*}
where the positive-definite Hessian matrix $A(t)\in\mathbb{R}^{n\times n}$ and vector $\vec{b}(t)\in\mathbb{R}^{n}$ are known. The aim is to solve the unknown dynamic vector $\vec{x}(t) \in\mathbb{R}^{n}$ in real-time \cite{FTZNNBook}. Next, an auxiliary function is defined as $f(\vec{x}(t), t)=\frac{1}{2}\vec{x}^{\text{T}}(t)A(t)\vec{x}(t)-\vec{b}^{\text{T}}(t)\vec{x}(t)$, and its gradient is formulated as $\nabla f(\vec{x}(t), t)=\partial f(\vec{x}(t), t)/\partial \vec{x}(t)=A(t)\vec{x}(t)-\vec{b}(t)$. The error function is defined to monitor the solution process:
\begin{linenomath*}
\begin{equation}\label{ErrD}
	\vec{e}(t) = A(t)\vec{x}(t)-\vec{b}(t).
\end{equation}
\end{linenomath*}
Then combining the evolution function of the OZNN \eqref{ZNNEF} and error function \eqref{ErrD}, the model expression is presented as $A(t)\vec{\dot x}(t)=-\dot A(t)\vec{x}(t)+\vec{\dot b}(t)-\gamma \Psi(A(t)\vec{x}(t)-\vec{b}(t))$, where $\dot A(t)$, $\vec{\dot x}(t)$, and $\vec{\dot b}(t)$ represent corresponding parameters's time derivative. Specifically, the optimal solution of the DQM problem \eqref{DQM} is found to be equivalent to when residual error generated by the error function \eqref{ErrD} converges to a predefined accuracy. Other solutions for dynamic tasks based on the ZNNs can be generated by a similar process with extra preprocess procedures, and representative problems are provided in Table \ref{PF}. Specifically, ZNNs are applied to solve dynamic nonlinear equation problems in the early stage. Then aided by the Lagrange multiplier method, ZNNs are extended to handle dynamic optimization problems with equality constrained \cite{JinSurvey}. After that, ZNNs are further developed to systematically solve the multi-inequality constraints dynamic optimization problems \cite{CDCLS}. Besides, extensive applications can be incorporated into this solution framework, such as robotic control and portal crane system control \cite{LFYConOpt}.
\par
Excepting representative problems summarized in Table \ref{PF}, ZNNs are extended for new issues recently. Matrix factorization is one of the emerging topics in ZNNs. First, considering the QR decomposition is demanded in factorization problems and lacking effective methods to deal with the dynamic cases. A dynamic complex-valued QR decomposition method is presented via ZNNs which is also capable of handling static situations \cite{QRDOne}. Also, there are other matrix factorization issues, e.g., LQ decomposition \cite{LQD}. Second, expanding the vector or matrix from a real-valued problem to a complex-valued, quaternion-valued, or tensor form problem is another hot spot in ZNNs \cite{Jiang}\cite{XLOne} \cite{WGC} \cite{Qua}. For example, a ZNN is presented with rigorous mathematical derivation investigation regarding convergence and robustness when solving the complex-valued quadratic programming \cite{Jiang}. On the aspect of the dynamic tensor problem, Mo $et~al.$ develop a finite-time ZNN for solving the tensor square root problem \cite{FTZNNEF}.

\begin{table}[t]
	\centering 
	\begin{tabular}[l]{@{}c l c}
	\toprule[2pt]
	Model &Evolution Function\\
	\toprule[1pt]
	OZNN & $\vec{\dot e}(t)=-\gamma\Psi(\vec{e}(t))$\\
	VPZNN & \multirow{2}*{$\vec{\dot e}(t)=-\mu(t)\Psi(\vec{e}(t))$}\\
	\cite{VPZNN} & \\
	NTZNN & \multirow{2}*{$\vec{\dot e}(t)=-\gamma \vec{e}(t)-\beta\int^t_0 \vec{e}(\delta)\text{d}\delta$}\\
	\cite{IEZNN} & \\
	FTZNN & \multirow{2}*{$\vec{\dot e}(t)=-\gamma \Psi\big{(}a_1 \vec{e}(t)+a_2 \vec{e}^{b/c}(t)\big{)}$}\\
	\cite{FTZNNEF} & \\
	Activated NTZNN & $\vec{\dot e}(t)=-\gamma \Psi_1 \big{(}\vec{e}(t)\big{)}$\\
	\cite{IEZNNOne}& $-\beta\Psi_2\Big{(}\vec{e}(t)+\gamma\int^t_0\Psi_1\big{(}\vec{e}(\delta)\big{)}\text{d}\delta\Big{)}$\\
	\toprule[2pt]
\end{tabular}
\caption{Comparisons among Different Evolution Formulas}
\label{EF}
\end{table}

\begin{table*}[t]
	\centering
	\begin{tabular}[l]{@{}l l l l}
	\toprule[2pt]
	Dynamic Problem & Problem Definition & Error Function \\
	\toprule[1pt]
	Linear system& $A(t)\vec{x}(t)=\vec{b}(t)$& $\vec{e}(t)=A(x)\vec{x}(t)-\vec{b}(t)$\\
	Stein matrix equation & $A(t)X(t)B(t)+X(t)=C(t)$ & $E(t)=A(t)X(t)B(t)+X(t)-C(t)$\\
	Nonlinear equation& $f(\vec{x}(t), t)=0$ & $\vec{e}(t)=\vec{g}(\vec{x},t)=\frac{\partial f(\vec{x}(t), t)}{\partial \vec{x}(t)}$\\
	Matrix square roots finding& $X^2(t)=A(t)$ & $E(t)=X^2(t)-A(t)$\\
	Matrix inversion& $A(t)X(t)=I$ & $E(t)=A(t)X(t)-I$\\
	\multirow{3}*{Quadratic programming$^{\dagger}$} &\multirow{2}*{$\min_{\vec{x}(t)} \frac{1}{2}\vec{x}^{\text{T}}(t)A(t)\vec{x}(t)+\vec{b}^{\text{T}}(t)\vec{x}(t)$} & \multirow{3}*{$\vec{e}(t)=\begin{bmatrix}
        	A(t)~D^\text{T}(t)\\
        	D(t)~~~\vec{0}~
    	\end{bmatrix}
    	\begin{bmatrix}
        	\vec{x}(t)\\
        	\vec{\rho}(t)
    	\end{bmatrix}-
    	\begin{bmatrix}
        	-\vec{b}(t)\\ 
        	\vec{c}(t)
    	\end{bmatrix}$} \\
	&\multirow{2}*{$~~\text{s.t.} ~~D(t)\vec{x}(t)=\vec{c}(t)$} \\
	\\
	\multirow{2}*{Linear equation and}& \multirow{3}*{$
	\left\{
	\begin{array}{ll}
		A(t)\vec{x}(t)=\vec{b}(t)\\
		C(t)\vec{x}(t)\leq \vec{d}(t)
	\end{array}
	\right.$} & \multirow{3}*{$\vec{e}(t)=\begin{bmatrix}
        	A(t)~~~\vec{0}\\
        	C(t)~D(t)
    	\end{bmatrix}
    	\begin{bmatrix}
        	\vec{x}(t)\\
        	\vec{y}(t)
    	\end{bmatrix}-
    	\begin{bmatrix}
        	-\vec{b}(t)\\
        	\vec{d}(t)
    	\end{bmatrix}$}\\
    	\multirow{2}*{inequality systems}\\
    	\\
	\multirow{3}*{Lyapunov equation}& \multirow{3}*{$A^{\text{T}}(t)X(t)+X(t)A(t)+B(t)=0$} &\multirow{2}*{$\vec{e}(t)=-(I\otimes A^{\text{T}}(t) + A^{\text{T}}(t)\otimes I)\text{vec}(X(t)))$}&\\ &&\multirow{2}*{$+\text{vec}(Q(t))$} \\
	\\
	\multirow{2}*{Sylvester equation} & \multirow{2}*{$A(t)X(t)-X(t)B(t)+C(t)=0$} &$\vec{e}(t)=(I\otimes A(t)-B^{\text{T}}(t)\otimes I)\text{vec}(X(t))$\\&&$+\text{vec}(C(t))$ \\
	Yang-Baxter-like equation& $X(t)A(t)X(t)=A(t)X(t)A(t)$ & $E(t)=X(t)A(t)X(t)-A(t)X(t)A(t)$\\
	\toprule[2pt]
\multicolumn{4}{l}{Symbol $\otimes$ represents the Kronecker product operation and $\text{vec}(\cdot)$ denotes the vectorization.}\\
\multicolumn{4}{l}{$^{\dagger}$ Parameter $\vec{\rho}(t)$ denotes the Lagrange-multiplier vector.}
\end{tabular}
\caption{Representative Dynamic Problems}
\label{PF}
\end{table*}

\section{Various Models}\label{VM}
In this section, the recent advance of each category of ZNNs is summarized and discussed in detail one by one.
\subsection{Noise-Tolerant Model}\label{NTM}
The stability of the solver for dynamic problems under perturbation influences is critical and directly determines the performance in practical applications. ZNNs are vulnerable to perturbations, which causes reliability concerns on these models owing to the potentially severe consequences. To this end, the noise-tolerant ZNN (NTZNN) is presented to guarantee the robustness of solutions in the presence of noise injected without any extra noise filter operations \cite{IEZNN}. The NTZNN is a milestone architecture in the history of ZNNs and has shown to possess inherent noise immune properties, thereby becoming the cornerstone of designing ZNNs \cite{SAAF}. The NTZNN utilizes the integration item to suppress noise disturbance, and the corresponding evolution formula is constructed as
\begin{linenomath*}
\begin{equation*}
	\vec{\dot e}(t)=-\gamma \vec{e}(t)-\beta\int^t_0 \vec{e}(\delta)\text{d}\delta,
\end{equation*}
\end{linenomath*}
where the scale parameters $\gamma>0$ and $\beta>0$. Noteworthy, the NTZNN is usually a benchmark framework to design the solution of dynamic problems rather than be used individually \cite{YJK} \cite{IEZNNOne}.

\subsection{Varying-Parameter Model}\label{VPM}
Previous ZNNs adopt the fixed scale parameter and have limitations that lead to slower convergence speed or require redundant manual fine-tuning. To address these inadequacies, the varying-parameter ZNN (VPZNN) is presented and furnished with monotonically increasing varying scale parameters for accelerating the convergence speed of the solving system \cite{VPZNNOne}. The main idea of the VPZNN is to use the following varying-parameter evolution formula to develop solutions:
\begin{linenomath*}
\begin{equation*}
	\vec{\dot e}(t)=-\mu(t)\Psi(\vec{e}(t)).
\end{equation*}
\end{linenomath*}
Benefitting employing the time-varying parameter $\mu(t)$, the VPZNNs are shown to possess super-exponential convergence and maintain the solution accuracy under noise perturbed without any extra structure such as an integral item or nonlinear activation functions \cite{VPZNN}. Following this, a power-type parameter VPZNN is designed to solve the noises perturbed quadratic minimization and quadratic programming problems, which is aided by mathematical derivation analyses to ensure robustness theoretically \cite{PTVP}. Besides, the VPZNN has been developed for various applications, such as Coronavirus diagnosis \cite{DL} and robot manipulators \cite{RKZNN}.

\subsection{Finite-Time Convergence Model}
The traditional ZNNs only consider fixed scale and basic activation functions such as linear or sigmoid-like functions, thus sharply reducing the convergence rate when solution error approaches zero. Besides, the scale parameters of the VPZNN increase monotonically with time, which sometimes leads to calculation overflow \cite{FTZNNBook}. To this end, inspired by the concept of the continuous autonomous system, the ZNN with finite-time convergence model (FTZNN) is presented \cite{FTZNNConcept}. As an improvement, the FTZNN has been widely researched and as a cornerstone in many practical applications since being proposed. Methods for realizing finite-time convergence can be roughly divided into three categories and suitable for different situations. The evolution formula-based scheme is designed as follows rather than the traditional scheme mentioned in the formula \eqref{ZNNEF} to achieve finite-time convergence:
\begin{linenomath*}
\begin{equation*}
	\vec{\dot e}(t)=-\gamma \Psi\big{(}a_1 \vec{e}(t)+a_2 \vec{e}^{b/c}(t)\big{)},
\end{equation*}
\end{linenomath*}
where the parameter $a_1>0$, $a_2>0$, and odd integer $b$ and $c$ satisfy $b>c>0$. On this basis, through embedding time-varying parameters, a new FTZNN with faster convergence is presented and comprehensively analyzed to resolve tensor square root finding \cite{FTZNNEF}. 
\par
On the other aspect, the activation function-based scheme is enabled finite-time convergence by utilizing nonlinear activation functions such as the sign-bi-power activation function \cite{FTZNNConcept}. Recently, several papers have been devoted to employing this scheme to exploit the FTZNN due to the scalability of the activation function-based scheme. After that, in order to handle the dynamic complex linear equation, the bounded nonlinear function is presented and investigated, which matches the real application scenes \cite{WGC}. Furthermore, several other works are devoted to solving time-varying problems or realizing applications with different finite-time convergence function realizations, such as dynamic complex-valued Lyapunov equation \cite{HYJ}. Additionally, relying on the modification scale parameter, the FTZNN is able to accelerate the convergence process further. On this principle, the FTZNN model with changing parameters is established for solving the complex linear matrix equation \cite{XLOne}. Theoretical results verify that this model has both global super-exponential and finite-time convergence performance. To solve the dynamic Moore-Penrose inverse problem with noises injected, a new evolution formula is presented and leads to two varying parameter FTZNN models, which theoretically can preserve finite-time convergence in the presence of noises perturbed environments \cite{FTZNNTZG}.

\subsection{Nonconvex Projection Model}
The activation function employed in ZNNs is usually designed as a monotonically increasing convex function which restricts the development of activation functions to further improve the convergence of solutions. Although many activation functions have been presented to improve performance, such as sign-bi-power or bipolar-sigmoid, they require relatively redundant formulations. To address these deficiencies, the nonconvex projection ZNN (NPZNN) is presented and applied to solve the time-varying matrix pseudo inversion for the first time \cite{NPZNN}, which breakthrough the convex limitations on activation function and attempt to exploit nonconvex function to accurate convergence of solutions. Specifically, the concept for realizing the nonconvex projection is defined as
\begin{linenomath*}
\begin{equation*}
	R_{\Lambda}(D)=\mathop{\arg\min}_{G\in \Lambda}\|G-D\|_{2},
\end{equation*}
\end{linenomath*}
with $0\in \Lambda$, $D$ and $\Lambda$ are two sets. Then, Jin $et~al.$ presented a dynamic matrix inversion solution method and analyzed the NPZNN from the perspective, which is the first work devoted to finding the connection between the NPZNN and control system \cite{Control}. Based on this principle, a model is presented to resolve the bound constrained underdetermined linear systems under a complex environment and applied to physically limited PUMA560 control implementation. In addition, the theoretical analysis and experiment results demonstrate its faster convergence \cite{LHY}. Thereafter, Xiao $et~al.$ developed a parallel computing method called the complex-valued ZNN realized by two novel complex-valued activated approaches and conducted a comprehensive analysis to investigate the convergence, stability, and robustness from the Lyapunov theory perspective \cite{XiaoXC}. After that, a modified NPZNN is proposed to handle linear equations \cite{SAAF}.

\subsection{Model Extension}
In addition to the vector or matrix form dynamic problems in the real-valued domain, the linear activated ZNN can be employed in dynamic complex-valued tasks without any extra preprocess or specific structures, even to the quaternion-valued version \cite{LSComplex}. Note that the nonlinear function activated ZNN to handle complex-valued or quaternion-valued problems \cite{Qua} only required aided by simple extension approaches:
\begin{itemize}
\item First method: $\Psi_{\text{cv}}(G+iH)=\Psi(G)+i\Psi(H)$
\item Second method: $\Psi_{\text{cv}}(G+iH)=\Psi(\Omega)\odot\exp(i\Gamma)$,
\end{itemize}
where the parameter $i$ is the imaginary unit, $G\in\mathbb{R}^{m\times n}$ and $H\in\mathbb{R}^{m\times n}$ denote the real and imaginary parts of the complex matrix, respectively, parameter matrices $\Omega\in\mathbb{R}^{m\times n}$ and $\Gamma\in(0,2\pi]^{m\times n}$ represent the element-wise module and argument of the complex matrix, separately, and $\odot$ is Hadamard product operation \cite{Jiang}. The abovementioned approaches allow nonlinear activation functions to be applied to boost the complex-valued ZNNs and can satisfy different tasks \cite{XiaoXC} \cite{XLOne} \cite{WGC}. Besides, tensors as a generalization of matrices, appear in many applications, while little research aims to resolve the dynamic tensor problems dedicatedly and efficiently. Consequently, as a powerful method for dealing with dynamic problems, ZNNs have been developed and investigated for sundry dynamic tensor problems \cite{tensorOne} \cite{FTZNNEF}. As mentioned in \cite{tensorOne}, a modified ZNN is presented and investigated based on the Lyapunov theory for solving the multi-linear system with $\mathcal{M}$-tensors. After that, Mo $et~al.$ \cite{FTZNNEF} further applied the ZNN to the dynamic tensor square root in finite-time.

\section{Model Discrete Methods}\label{MDM}
The continuous-time ZNNs hypothesis is that the sample gaps are arbitrarily variable without communication time consumption. Remarkably, this hypothesis is an ideal precondition that cannot be simulated on a computer or directly implemented on a hardware circuit. Therefore, the continuous-time ZNNs must be discretized as a discrete-time version for being performed in realistic environments. To this end, this section highlights the prominent features of ZNNs from a discrete perspective and then provides the current common or famous discretization methods of ZNNs.
\subsection{Online Solution}
The features of the models or algorithms for handling time-varying issues are summarized in this subsection which is intrinsically different from the time-invariant problems solution scheme. Then, based on these characteristics, discrete-time ZNNs are introduced clearly and accurately.
\subsubsection{Predict Manner Computation and Real-Time Solving}
Methods used for solving time-varying problems are required to generate the future (next-step $x_{k+1}$) data only exploiting the present $x_{k}$ and previous $x_{k-1}$ information \cite{Shi}. In other words, the calculated solution $x_{k+1}$ of the aim problem at the time instant $t_{k+1}$ should be calculated before the time instant $t_{k+1}$. Then, at the time instant $t_{k+1}$, the solution method employs data $x_{k+1}$, which has been generated already. From this perspective, finding a reliable approach for realizing discrete-time methods without future information involved is a vital issue \cite{TTDZNN}. Computation consumed in each step is another fundamental issue in designing the discrete-time model. Specifically, many sophisticated numerical discretization approaches implement pre-calculations before updating one step, which requires extra computation time. Nonetheless, time resource is limited in the time-varying problems solution procedure, which often makes impossible directly transfer general numerical discretization approaches to time-varying problem solving methods.

\subsection{Discrete Methods}
Three common and widely used discrete ZNNs methods are summarized in this subsection.
\subsubsection{Euler Difference Methods}
Euler forward or backward differences are the common and basic discrete methods of ZNNs. Their mathematical representation is defined as $\vec{\dot x}_k=(\vec{x}_{k+1}-\vec{x}_{k})/\eta+O(\eta)$ for the forward difference, and $\vec{\dot x}_k=(\vec{x}_{k}-\vec{x}_{k-1})/\eta+O(\eta)$ for the backward difference where $\eta$ denotes the sample gap and the larger value results in more efficient computation and less precision. As an example, in light of the definition of the Euler forward difference and OZNN for solving the DQM problems \eqref{DQM}, the corresponding discrete-time version with forward difference adopted is formulated as
\begin{linenomath*}
\begin{eqnarray*}
	\begin{split}
		\vec{x}_{k+1}=&\vec{x}_k-A^{\dagger}_k\Big{(}(A_{k}-A_{k-1})\vec{x}_k-(\vec{b}_k-\vec{b}_{k-1})\\
		&+\eta\gamma(A_k\vec{x}_k-\vec{b}_k)\Big{)}.
	\end{split}
\end{eqnarray*}
\end{linenomath*}
The other models such as NTZNN and VPZNN mentioned previously are produced as the same procedure. For example, discrete-time ZNNs are developed to resolve perturbed complex quadratic programming \cite{QYM} and continuum robots control scheme \cite{NT} by using this approach. Noteworthily, as presented in \cite{NT}, Euler difference methods can be extended to a three-step version as
\begin{linenomath*}
\begin{equation*}
	\vec{\dot x}_k=(2\vec{x}_{k+1}-3\vec{x}_{k}+2\vec{x}_{k-1}-\vec{x}_{k-1})/(2\eta)+O(\eta^2),
\end{equation*}
\end{linenomath*}
or even higher multi-step difference versions to further improve the precision of the solution.

\subsubsection{Taylor-Type Differentiation Methods}
In order to improve computational precision and eliminate the divergence phenomenon, Zhang $et~al.$ \cite{TTDZNN} present the Taylor-type differentiation method, and the mathematical representation is furnished as
\begin{linenomath*}
\begin{equation*}
	\dot y(x_{t+1})=\frac{2y(x_{k+1})-3y(x_{k})+2y(x_{k-1})-y(x_{k-2})}{2\eta}+\Delta,
\end{equation*}
\end{linenomath*}
where $\Delta=O(\eta^2)$. Consequently, the Taylor-type discrete-time OZNN for solving the DQM problem \eqref{DQM} is formulated
\begin{linenomath*}
\begin{eqnarray*}
	\begin{split}
		\vec{x}_{k+1}=&\frac{3}{2}\vec{x}_{k}-\vec{x}_{k-1}+\frac{1}{2}\vec{x}_{k-2}\\
		&-A^{-1}_{k}\Big{(}\eta\dot A_{k}\vec{x}_{k}+\eta\gamma(A_k \vec{x}_k-\vec{b}_k)-\eta \vec{\dot b}_k\Big{)}.
	\end{split}
\end{eqnarray*}
\end{linenomath*}
Currently, many discrete-time ZNNs are developed to deal with dynamic problems based on this role. In \cite{TDZNNSR}, a Taylor-type discrete-time ZNN is developed to handle the dynamic quadratic programming and theoretically analyze the influence of step size on the proposed model, which can further lead to the optimal setting. Besides, Taylor-type differentiation methods show potential and scalability, thereby being widely developed \cite{YMOne}.
\subsubsection{Runge-Kutta Methods}
Runge-Kutta (RK) method is a famous ordinary differential equation iterative method, including implicit and explicit fashion, which can be applied to discrete ZNNs. As a classical RK method, the fourth-order RK method is developed in \cite{RKZNN} to achieve a redundant robot control approach. Furthermore, the `ode45' function in MATLAB, which is realized by the fourth and fifth-order RK method with variable-step, is widely adopted to simulate the continuous-time ZNNs.

\begin{table*}[t]
	\centering
	\begin{tabular}[l]{@{}c c c c c |c c c}
	\toprule[2pt]
	\multirow{2}*{Type}&\multirow{2}*{Reference} &Global &Convergence &Finite-Time &\multicolumn{3}{|c}{Robustness Analysis}\\
	&&Convergence&Speed &Convergence &Constant &Linear &Bounded Random \\
	\toprule[1pt]
	\multirow{2}*{NTZNN}&\cite{IEZNN} &\cmark &\cmark &- &\cmark &\cmark &\cmark\\
	&\cite{IEZNNOne}$^{\dagger}$ &\cmark &- &- &\cmark &- &\cmark\\
	\multirow{1}*{VPZNN}&\cite{VPZNN} &\cmark &\cmark &- &\cmark &\cmark &-\\
	\multirow{2}*{FTZNN}&\cite{FTZNNBook} &\cmark &\cmark &\cmark &\cmark &\cmark &\cmark\\
	&\cite{FTZNNEF} &\cmark &\cmark &\cmark &- &- &-\\
	\multirow{1}*{NPZNN}&\cite{SAAF} &\cmark &- &- &\cmark &\cmark &\cmark\\
	\multirow{2}*{Complex}&\cite{Jiang} &\cmark &- &- &\cmark &\cmark &-\\
	&\cite{WGC} &\cmark &- &\cmark &- &- &-\\
	\toprule[2pt]
\end{tabular}
\caption{Theoretical Analysis of ZNNs}
\label{ZNNPerAna}
\end{table*}

\section{Theoretical Analyses}\label{TAsection}
This section revisits the analysis approaches of ZNNs. Theoretical analysis is an important part of ZNNs research area. Leveraging the rigorous mathematical derivation to investigate the performance of the specific ZNN before being deployed in practical applications is an indispensable procedure \cite{JinSurvey}. The connection between continuous-time and discrete-time ZNNs is that the former can be transformed into the latter by employing numerical differential approaches mentioned in Section \ref{MDM}. Further, the continuous-time ZNNs are usually analyzed by an ordinary differential equation \cite{XLOne} \cite{IEZNNOne}, Laplace transforms \cite{IEZNN} \cite{Liufu}, and Lyapunov theory \cite{LHY} while discrete-time ZNNs are analyzed as a numerical algorithm. Some representative papers regarding convergence, stability, and precision cross different ZNNs are provided in Table \ref{ZNNPerAna} for a supplement.

\subsection{Convergence Analysis}
Convergence is a widely investigated property of ZNNs. Specifically, global convergence is a basic premise for the ZNN global approaches to the theoretical solution. Besides, the convergence speed shows that the speed of the calculated solution approaches the theoretical solution \cite{VPZNN}. The Lyapunov theory is a broadly adopted analysis framework for global convergence \cite{XiaoXC}, which should define a candidate function formally as $L(t)=\vec{e}^{\text{T}}(t)\vec{e}(t)/2$, then lead to the corresponding time derivative $\dot L(t)$. Afterward, the global convergence is judged according to the definition of function $L(t)$ and $\dot L(t)$ \cite{Jiang}. Convergence speed performance relies on the ordinary differential equation analysis \cite{XLOne}. For instance, the evolution function of OZNN \eqref{ZNNEF} is formulated as $\vec{e}(t)=\vec{e}(0)\exp(-\gamma t)$ by using the general solutions of the ordinary differential equation then leads to the convergence speed \cite{VPZNNOne}. Note that a similar process can generate the other ZNNs' analysis methods and conclusions by replacing evolution functions, even for complex-valued ZNNs \cite{LHY} \cite{YJK}.
\subsection{Robustness Analysis}
Noise perturbation serves as an essential surrogate to evaluate the robustness of ZNNs before models are deployed \cite{WGC}. The robustness of the existing ZNNs hinges on the noise suppression ability. Further, robustness analyzes are leveraged to evaluate the stability of the noise perturbations injected ZNNs for directing improvement methods \cite{SAAF}. Improving robustness can be achieved by employing the noise-tolerant evolution formula structure as Subsection \ref{NTM}, time-varying parameter as Subsection \ref{VPM}, and modified activation functions \cite{Jiang}. Specifically, the robustness analysis under constant, linear, and bounded random noises of the NTZNN using Laplace transform methods are presented in \cite{IEZNN}. After that, the activated NTZNN model with two different implementation approaches is provided in \cite{SZB} and \cite{SAAF}. Meanwhile, Zhang $et~al.$ theoretically prove that VPZNNs possess the ability of noise suppression \cite{PTVP}.

\section{Applications}\label{App}
As a flexible and efficient model with systematic theoretical support, ZNNs have shown great potential and have been used in various scenarios. Recently, in addition to the continuous progress in the field of automatic control, applying ZNNs to economics and image processing has become a new research branch. Hence, the revisit covered the recent advance of applications in the ZNNs field is provided in this section.
\subsection{Automatic Control}
Since ZNNs were proposed, they have rapid progress and manifested state-of-the-art performance in the field of robot control and multi-objective collaboration. Thereby many works focus on these tasks, and they have been summarized previously \cite{RobotOne}. Expressly, in \cite{PNNJin}, a systematic revisit of robotics implementation schemes regarding redundant robot manipulators, mobile robotics, and multiple cooperative control is provided, and discuss the advance of projection neural networks on these topics. In the other respect, Zhang $et~al.$ summarized VPZNNs' progress on the scenes of similar applications and supplied the VPZNNs' achievements on unmanned aerial vehicles and investment problems \cite{VPZNNOne}. Recently, the Zhang-gradient method has been presented and investigated in detail for developing a controller supplied by abundant example verification, which is shown to possess high availability for many control system applications \cite{ZGC}. In addition to the above systematically summarized literature, recent papers further promote progress in these fields. First, from the perspective of game theory, Zhang $et~al.$ leverage the NTZNN with the activation function employed to develop a distributed method for improving the manipulability of manipulators in a group \cite{ZJZOne}. This work provides a new paradigm in multi-manipulator control. Second, a super twisting algorithm aided ZNN based control scheme named STZNN model is developed for mobile robot manipulators, which points out the intrinsic relationship of ZNNs and sliding mode control. The STZNN model can eliminate the infinitely time convergence phenomenon and improve robustness \cite{CDCRobot}.

\subsection{Mobile Object Positioning}
ZNNs enable to process the dynamic problem or fit the time-varying system in a real-time fashion, thereby showing competitive performance when realizing mobile object positioning schemes \cite{YJK}. Current research is committed to using the ZNNs to replace conventional methods, such as pseudoinverse or gradient descent, to establish the positioning scheme. Benefitting from the features of ZNNs, these approaches can eliminate the lagging error and ensure the robustness of real-time solutions \cite{QRDOne}.
\subsubsection{Angle-of-Arrival Based Scheme}
The principle of the angle-of-arrival (AoA) positioning scheme utilizes the geometric relationship between the base station and the moving target to estimate the real-time location \cite{Huang}, and the mathematical description of a three--dimensional AoA positioning scheme is defined as a dynamic linear system. From this principle, several ZNN models are presented and used to develop effective positioning techniques \cite{HYJ} \cite{YM} \cite{AoAThree}. Aiding by existing nonlinear activation functions, a new nonlinear activation function is presented and exploited to realize a ZNN model for reducing noise influence when applied to positioning schemes \cite{HYJ}. Thereafter, to improve the solution precision, explicit linear left-and-right 5-step formulas are furnished in the ZNN model and used to realize the AoA positioning scheme \cite{YM}. Besides, a high order ZNN model is presented in \cite{AoAThree} to investigate the capacity of the proposed model.

\subsubsection{Time Delay of Arrival Based Scheme}
The time delay of arrival (TDOA) positioning scheme is modeled based on the relationship between the time delay of targets to different observers and the transmission speed of the signal in the medium \cite{YJK}. The mathematical expression of the three-dimensional version of the TDOA positioning scheme \cite{AoA} is defined as
\begin{linenomath*}
\begin{equation}\label{TDOA}
	\begin{bmatrix}
       a_2 & b_2 & c_2 & vd_{2}(t)\\
      \vdots & \vdots & \vdots & \vdots \\
      a_m & b_m & c_m & vd_{m}(t)
    \end{bmatrix}
	\begin{bmatrix}
       x(t)\\
       y(t)\\
       z(t)\\
       r_1(t)
    \end{bmatrix}
	=
        \begin{bmatrix}
        vd_{2}(t)-g_{2}(t)\\
        \vdots\\
        vd_{m}(t)-g_{m}(t)
        \end{bmatrix},
\end{equation}
\end{linenomath*}
where $(a_i,b_i,c_i)$ represents the position of $i$th ($i\in\{2,\cdots,m\}$) observer, and $(x(t),y(t),z(t))$ denotes the target position. The parameter $r_i(t)$ signifies the Euclidean distance between the target and the first observer, $d_i(t)$ represents the time delay between the $i$th and the first observer, and $g_{i}(t)$ is constructed as $g_i(t)=h^2_i(t)+2r(t)h_i(t)+2x_ix(t)+2y_iy(t)+2z_iz(t)$, where the $h_i(t)=r_i(t)-r_1(t)$. Based on the TDOA, a noise-suppressing ZNN model is presented to realize an acoustic source localization method, which is an earlier attempt to apply ZNNs to this task \cite{YJK}.

\subsection{Image Processing}
ZNNs are shown to possess high hardware implementability and can handle problems in a zero-error manner, satisfying the exact solution requirement in image processing problems. Therefore, despite ZNNs being proposed for dynamic problems inherently, they can also achieve transfer to scenarios like image reconstruction \cite{ImgNoiseOne} or target detection \cite{Jiang} \cite{SAAF} \cite{Liufu}. 
\subsubsection{Image Denoise}
A ZNN model with predefined-time convergence is presented and applied to realize the multi-sensors image denoise scheme aided by the data fusion method \cite{ImgNoiseOne}. The image denoise scheme converts the potential problem to a constrained optimization problem and then exploits the proposed ZNN model to solve it. Hereafter, two models are presented to address image denoise problem, with robustness enhanced and incorporating fuzzy parameters, respectively \cite{FTZNNBook}.
\subsubsection{Target Detection}
Recently, some research efforts have been made to develop ZNNs for implementing the image target detection scheme. In \cite{Jiang}, an easy but effective remote sensing image targets detection scheme is designed, which is the first attempt to apply the ZNNs to achieve image targets detection. An approach based on the ZNN is also derived with the same potential optimization problem but employed for the image enhancement task \cite{SAAF}. After that, an algorithm is presented with lower computational complexity and more stable performance and performed in more different or complicated scenes \cite{Liufu}.

\subsection{Other Applications}
Beyond the above applications, several works have recently committed to exploiting the ZNN or their main idea, to develop or improve deep learning algorithms \cite{DL} \cite{DLOne}. First, to recognize the coronavirus disease 2019 from the X-ray chest radiography images in an automatic manner, the deep ensemble dynamic learning network is presented based on the ZNN evolution function. This method achieves good performance on the open image dataset \cite{DL}. Second, the acceleration optimizers are proposed based on ZNNs for accelerating the convergence and reducing loss when training the deep neural network, which is the first attempt to integrate ZNN into the deep neural network \cite{DLOne}. Markowitz mean-variance portfolio selection is an extensively used and effective investment method. 

\section{Discussions}
ZNNs emerge as a powerful tool for dynamic problems and have shown rapid development over the past twenty years. Meanwhile, the corresponding comprehensive analysis theories regarding the complexity, convergence, and robustness of different models are refined to guide the advancement of ZNNs. Therefore, ZNNs have become a systematic and baseline solution for dynamic tasks. Nonetheless, there are still unanswered problems or possibly advanced methods that have yet to be considered. Consequently, we summarize these under-researched problems and provide an open perspective to looking forward to the future of ZNNs.
\begin{itemize}
\item From the perspective of optimization, the integration enhanced method employed in the NTZNN can be deemed as the heavy-ball method, which exploits historicafl information to improve performance. However, to the best of our knowledge, no paper uses Nesterov’s accelerated method to improve the convergence of ZNNs further. Hence, developing methods to improve the performance by exploiting the Nesterov method is valuable.
\item The inherent relationship between ZNNs and the Newton-Raphson method has been reported. Inspired by this point, we can extend the design in ZNNs to advanced algorithms such as proximal gradient or mirror descent methods for exploring solutions to complex dynamic problems (e.g., non-differentiable or composite optimization seems feasible).
\item The efficient discrete calculation method is a basic and critical issue in ZNNs, which is the premise of guaranteeing a real-time solution. Although discrete methods to perform ZNNs exist, using the nonlinear activation function or time-varying scale parameter sometimes requires more computing resources, leading to more computation time during each iteration, thereby failing the real-time solution.
\item Currently, ZNNs have been applied to many practical applications covering automatic control, image processing, economics, etc. Nevertheless, there still exists many application scenarios worth exploring how to use ZNNs to develop solutions for improving performance, such as nonconvex optimization and deep learning.
\end{itemize}

\section{Conclusion}
This paper summarizes the advance of zeroing neural networks (ZNNs), covering methodology, theory, and applications. We hope this work can help the community better understand and learn about the current development of ZNNs.


\bibliographystyle{named}
\bibliography{Manuscript}

\end{document}